\documentclass{article}
\usepackage{float}
\usepackage{arxiv}
\usepackage[utf8]{inputenc} % allow utf-8 input
\usepackage[T1]{fontenc}    % use 8-bit T1 fonts
\usepackage{hyperref}       % hyperlinks
\usepackage{url}            % simple URL typesetting
\usepackage{booktabs}       % professional-quality tables
\usepackage{amsfonts}       % blackboard math symbols
\usepackage{nicefrac}       % compact symbols for 1/2, etc.
\usepackage{microtype}      % microtypography
\usepackage{graphicx}
\usepackage{doi}
\usepackage{amsmath}
\usepackage{algorithm}
\usepackage{algpseudocode}
\usepackage{cite}

\title{Understanding Malware Propagation Dynamics through Scientific Machine Learning}

\author{
    Karthik Pappu\\
    Dakota State University\\
    \texttt{karthik.pappu@trojans.dsu.edu}\\
    \And
    Prathamesh Dinesh Joshi\\
    Vizuara AI Labs\\
    \texttt{prathamesh@vizuara.com}\\
    \And
    Raj Abhijit Dandekar\\
    Vizuara AI Labs\\
    \texttt{raj@vizuara.com}\\
    \And
    Rajat Dandekar\\
    Vizuara AI Labs\\
    \texttt{rajatdandekar@vizuara.com}\\
    \And
    Sreedath Panat\\
    Vizuara AI Labs\\
    \texttt{sreedath@vizuara.com}\\
}

\renewcommand{\shorttitle}{Malware Dynamics via SciML}

\hypersetup{
pdftitle={Understanding Malware Mutation Dynamics through Scientific Machine Learning},
pdfsubject={Computer Science, Machine Learning, Cybersecurity},
pdfauthor={Karthik Pappu},
pdfkeywords={Malware dynamics, Mutations, Scientific Machine Learning, ODE, Neural ODE, Universal Differential Equations},
}

\begin{document}
\maketitle

\begin{abstract}
Accurately modeling malware propagation is essential for designing effective cybersecurity defenses, particularly against adaptive threats that evolve in real time. While traditional epidemiological models and recent neural approaches offer useful foundations, they often fail to fully capture the nonlinear feedback mechanisms present in real-world networks. In this work, we apply scientific machine learning to malware modeling by evaluating three approaches: classical Ordinary Differential Equations (ODEs), Universal Differential Equations (UDEs), and Neural ODEs. Using data from the Code Red worm outbreak, we show that the UDE approach substantially reduces prediction error compared to both traditional and neural baselines by 44\%, while preserving interpretability. We introduce a symbolic recovery method that transforms the learned neural feedback into explicit mathematical expressions, revealing suppression mechanisms such as network saturation, security response, and malware variant evolution. Our results demonstrate that hybrid physics-informed models can outperform both purely analytical and purely neural approaches, offering improved predictive accuracy and deeper insight into the dynamics of malware spread. These findings support the development of early warning systems, efficient outbreak response strategies, and targeted cyber defense interventions.
\end{abstract}

\keywords{Malware Propagation \and Malware Dynamics \and Scientific Machine Learning \and Universal Differential Equations \and Neural ODEs \and Cybersecurity}

\section{Introduction}

The rapid evolution of malware poses unprecedented threats to global cybersecurity infrastructure, with economic damages exceeding hundreds of billions of dollars annually \cite{afaq2023critical}. Contemporary malware exhibits increasingly sophisticated behaviors, including real-time adaptation, evasion techniques, and coordinated propagation strategies that challenge traditional modeling frameworks \cite{ormeir2019dynamic, bilge2012before}. Advanced persistent threats (APTs) and zero-day malware now represent particularly challenging scenarios, employing intelligence-driven targeting and stealthy propagation mechanisms that can remain undetected for extended periods while continuing to spread \cite{hernandez2019security}. These sophisticated attacks utilize decision-based targeting, where malware determines whether to actively exploit a compromised device or use it merely as a carrier for further propagation, fundamentally altering traditional infection dynamics. Understanding and predicting these complex dynamics is crucial for developing effective defense mechanisms, early warning systems, and response strategies that can adapt to emerging threats.

Traditional mathematical models of malware propagation draw inspiration from epidemiological frameworks, adapting SIR-type models originally developed for biological disease spread \cite{kermack1927contribution, hethcote2000mathematics} to cyber environments \cite{cohen1987computer, kephart1991directed, zou2002code}. While these approaches provide valuable baseline insights, they rely on simplified assumptions of uniform network behavior and fixed system parameters that frequently fail to capture the complex, non-linear behaviors present in real-world network environments \cite{pastor2001dynamical, zhuravel2024stochastic}. Critical phenomena such as network topology effects, traffic congestion, dynamic security countermeasures \cite{mirkovic2004taxonomy}, and adaptive malware techniques remain challenging to model with conventional mathematical approaches, resulting in suboptimal predictions during early outbreak phases when accurate forecasting is most essential \cite{staniford2002how, ganesh2005effect}.

The core challenge lies in the limitations of both traditional and modern approaches. Conventional epidemic models lack the flexibility to capture the complex, non-linear dynamics of modern malware spread. Meanwhile, data-driven machine learning approaches often require large amounts of labeled data (typically unavailable during emerging outbreaks) and tend to prioritize prediction accuracy at the expense of interpretability \cite{goodfellow2016deep}. Moreover, cybersecurity data is frequently characterized by irregular sampling intervals, measurement noise, and incomplete observations, creating additional challenges for both traditional mathematical models and neural network approaches \cite{goyal2022neural}. Zero-day malware detection further compounds these difficulties by providing minimal training samples for emerging threat variants, while the spatial-temporal nature of network-based propagation introduces reaction-diffusion dynamics that require sophisticated mathematical frameworks to capture effectively \cite{du2018partial}. Although recent efforts have advanced from early theoretical models \cite{cohen1987computer} to sophisticated network-based frameworks \cite{yu2015malware, pastor2015epidemic}, these still rely on fixed mathematical structures that struggle to adapt to rapidly evolving threats.

Scientific Machine Learning (SciML) offers a promising solution by integrating physics-based modeling with data-driven techniques \cite{karniadakis2021physics}. This approach combines the interpretability of mechanistic models with the flexibility of neural networks, enabling systems to capture both fundamental dynamics and emergent behaviors that purely analytical models cannot represent \cite{chen2018neural, rackauckas2020universal}. Recent advances in physics-informed neural networks have demonstrated success in handling irregular and noisy data across diverse domains, from vehicle platoon security \cite{vyas2023physics} to automated risk assessment systems \cite{wang2020automatic}, suggesting their potential applicability to cybersecurity challenges.

Despite these advances, current malware models continue to face difficulties in adapting to evolving network dynamics while preserving interpretability. Critical gaps remain in handling the stochastic, spatially distributed nature of modern malware propagation, managing uncertainty in real-time operational environments, and processing irregular, noisy observational data commonly encountered in cybersecurity applications. Contemporary approaches to malware detection increasingly rely on sophisticated techniques, such as graph neural networks \cite{bilot2024survey}, multi-loss architectures for feature learning \cite{li2021multi}, and federated learning for distributed scenarios \cite{purkayastha2023android}. However, these methods often sacrifice the mechanistic insights essential for understanding propagation dynamics. It represents a critical gap in cybersecurity, where both accurate forecasting and mechanistic insight are essential.

In this paper, we make the following contributions:

\begin{enumerate}
    \item We develop and evaluate three complementary modeling approaches (Ordinary Differential Equations (ODEs), Universal Differential Equations (UDEs), and Neural ODEs for modeling malware propagation dynamics using real-world Code Red worm data.
    
    \item We demonstrate that the UDE approach consistently outperforms both traditional ODEs and Neural ODEs across multiple dimensions, achieving a 44\% reduction in prediction error while maintaining interpretability.
    
    \item  We apply symbolic recovery to interpret the neural network component, revealing its role as a suppression mechanism that corresponds to observable cybersecurity phenomena.
    
    \item We map recovered mathematical terms to specific cybersecurity phenomena, revealing how network saturation, security response mechanisms, and variant evolution create suppression and amplification effects that govern real-world malware propagation.
    
    \item We provide a comprehensive analysis of each model's performance under varying conditions, including limited training data, measurement noise, and different forecasting horizons, offering practical guidance for model selection.
\end{enumerate}

Our findings reveal that malware propagation in digital networks exhibits fundamentally different dynamics than traditional epidemic models suggest, with network infrastructure limitations and dynamic security responses creating natural limiting factors that conventional frameworks fail to capture \cite{staniford2002how, zou2003monitoring}. These insights have immediate applications for cybersecurity practitioners seeking more accurate forecasting tools and for policymakers developing evidence-based cyber defense strategies.

\section{Related Work}
Mathematical modeling of malware propagation has evolved from classical epidemiological frameworks to modern hybrid approaches. Foundational work by Cohen \cite{cohen1987computer}, and Kephart and White \cite{kephart1991directed} adapted SIR models from epidemiology to digital threats. Zou et al.\ \cite{zou2002code} validated these methods using empirical Code Red worm data, while later studies introduced compartmental extensions for more nuanced infection states.

Network-aware models addressed the impact of topology on outbreak dynamics. Pastor-Satorras and Vespignani \cite{pastor2001dynamical} established that scale-free networks fundamentally alter epidemic thresholds, and Chernikova et al.\ \cite{chernikova2023modeling} demonstrated these principles with real-world analysis of self-propagating malware like WannaCry using enhanced epidemiological models. Infrastructure constraints also act as natural suppressors. Staniford et al.\ \cite{staniford2002how} highlighted bandwidth saturation and congestion effects during Code Red, while Ganesh et al.\ \cite{ganesh2005effect} formalized bottlenecks arising from network topology, factors absent from classical epidemic models.

Despite these advances, key limitations persist. Empirical studies show that SIR-style models achieve only moderate accuracy for malware forecasting \cite{ginters2023usability}, while deterministic ODEs often underestimate real-world uncertainty and stochasticity \cite{zhuravel2024stochastic}. Most ODE models assume homogeneous mixing, making them ill-suited for adaptive attacker behavior or dynamic defense responses. While recent extensions include agent-based models and stochastic differential equations, robust real-time forecasting and uncertainty quantification remain largely unresolved.

Modern malware poses new challenges due to zero-day attacks, polymorphism, and lateral movement techniques. Contemporary approaches have extended beyond traditional models to address these sophisticated threats. Du et al.\ \cite{du2018partial} developed spatial-temporal models using partial differential equations with mixed delays. In contrast, Hernández Guillén et al.\ \cite{hernandez2019security} introduced SCIRAS compartmental models specifically for advanced persistent threats that distinguish between carrier and targeted devices. Recent advances in graph neural networks show promise for malware propagation prediction \cite{li2023malware}. Modern detection approaches increasingly leverage graph representation learning \cite{bilot2024survey} and federated architectures \cite{purkayastha2023android}. While deep learning approaches have advanced static and dynamic malware detection \cite{ormeir2019dynamic}, their application to propagation modeling faces challenges in balancing predictive accuracy with interpretability. Pure neural approaches often favor performance over mechanistic understanding, which can limit their effectiveness for real-time defense strategies and scenario analysis.

Scientific machine learning offers a promising integration of mechanistic and neural modeling. Neural ODEs \cite{chen2018neural} parameterize continuous-time dynamics directly with neural networks, while UDEs embed neural components within physical models for greater flexibility and interpretability \cite{rackauckas2020universal}. Physics-informed machine learning approaches have shown particular promise for reliability and safety applications \cite{xu2023physics}, though their application to cybersecurity propagation modeling remains underexplored. Symbolic regression further enhances interpretability by extracting closed-form expressions from learned neural terms \cite{brunton2016discovering, schmidt2009distilling}, although practical applications in cybersecurity are still in their infancy.

Deployment remains challenging due to partial observability, noisy measurements, and the use of adversarial adaptation strategies. There is also growing interest in cross-domain modeling, where methods from epidemic theory and social contagion are adapted to enhance robustness and generalization \cite{pastor2015epidemic}.

Despite these advances, no comprehensive comparison exists between classical ODEs, Neural ODEs, and UDEs for malware propagation using real-world outbreak data. Our work addresses this gap by systematically evaluating scientific machine-learning models on dynamic host-based malware data and introducing symbolic recovery techniques for interpretability. We focus specifically on outbreak dynamics, leaving static file-based detection to previous reviews \cite{ormeir2019dynamic}.

\section{Modeling Methodology}

This section presents our data-driven modeling approach for malware dynamics. We begin with preprocessing infection data into a continuous-time representation, then progressively explore three modeling frameworks with increasing flexibility: mechanistic ordinary differential equations (ODEs), hybrid Universal Differential Equations (UDEs), and fully data-driven Neural ODEs.

\subsection{Data Preprocessing Pipeline}

We use the publicly available Code Red worm dataset provided by CAIDA \cite{caida2001codered}, which captures darknet telescope observations of scan activity associated with the Code Red worm outbreak. Each record in the raw tab-separated dataset includes seven fields: start and end times of scanning activity (Unix timestamps), source IP addresses, top-level domain, country, geographic coordinates (latitude/longitude), and Autonomous System (AS) metadata. These scans serve as proxies for worm infection attempts, providing temporal dynamics data suitable for mathematical modeling.

To convert this discrete event-based data into a form suitable for continuous-time modeling, we apply the following three-step preprocessing pipeline:

\begin{enumerate}
  \item \textbf{Temporal Binning}: Raw Unix timestamps are converted to absolute datetime format and grouped into uniform 30-minute intervals using pandas date range functionality. We selected 30-minute intervals to balance temporal resolution with statistical stability, providing sufficient data points per bin while capturing the dynamic nature of worm propagation. Each bin $i$ aggregates the number of scan events observed during that interval, yielding a discrete-time infection intensity signal $I_i$. Comment lines beginning with '\#' in the original data are automatically filtered during this process.
  
  \item \textbf{Smoothing}: To reduce high-frequency fluctuations in the binned intensity data and improve numerical stability for differential equation integration, we apply a 3-point moving average filter:
    \[
    \tilde{I}_i = \begin{cases}
    \frac{1}{2}(I_1 + I_2) & \text{if } i = 1 \\
    \frac{1}{3}(I_{i-1} + I_i + I_{i+1}) & \text{if } 1 < i < n \\
    \frac{1}{2}(I_{n-1} + I_n) & \text{if } i = n
    \end{cases}
    \]
    This bounded smoothing operation preserves endpoint values while stabilizing the signal for numerical integration. The smoothing operation reduces noise while maintaining the overall temporal structure of the outbreak progression. This preprocessing approach addresses the challenges of irregular sampling and noise \cite{goyal2022neural} commonly encountered in cybersecurity data.
    
  \item \textbf{Temporal Interpolation}: A continuous-time infection intensity function $\eta(t)$ is constructed via linear interpolation of the smoothed intensity signal using scipy's interpolation methods. This enables evaluation at arbitrary time points during differential equation integration and model simulation, providing the external forcing term required for our mathematical models.
\end{enumerate}

This preprocessing results in a continuous infection intensity function $\eta(t)$ suitable for integration within differential equation models.

\subsection{Classical ODE Model for Malware Dynamics}

We begin with a classical epidemic-style model formalized as an ordinary differential equation (ODE), designed to describe the nonlinear dynamics of malware spread in real-world networks. Grounded in foundational models of computer virus propagation \cite{kephart1991directed} and complex network epidemiology \cite{pastor2001dynamical}, this formulation serves as an interpretable baseline that incorporates key features such as logistic growth, external forcing, suppression effects, and adaptive feedback mechanisms \cite{zou2002code, fang2020statistical}.

To capture these dynamics in a cybersecurity context, we model the time evolution of the malware infection intensity \( M(t) \) using the following nonlinear ordinary differential equation:

\begin{equation}
\frac{dM}{dt} = \underbrace{\alpha(t) M \left(1 - \frac{M}{K}\right)}_{\text{logistic growth}} + \underbrace{\eta(t)}_{\text{external forcing}} - \underbrace{\beta M^2}_{\text{quadratic suppression}} + \underbrace{\kappa M \log(1 + M)}_{\text{adaptive feedback}}
\label{eq:malware-ode}
\end{equation}

Each term represents a specific infection mechanism:

\begin{itemize}
    \item \textbf{Logistic growth} models intrinsic malware propagation with resource constraints, where the infection rate decays exponentially over time: $\alpha(t) = \alpha_0 \exp(-p_{\text{decay}} \cdot t / t_{\max})$
    
    \item \textbf{External forcing} incorporates the empirical infection intensity through the interpolated function $\eta(t)$
    
    \item \textbf{Quadratic suppression} captures density-dependent effects like network congestion and resource exhaustion
    
    \item \textbf{Adaptive feedback} introduces positive reinforcement representing malware propagation dynamics. 
\end{itemize}

\subsubsection{Parameter Specification and Optimization}

Through empirical fitting to the Code Red data, we identified the following optimal parameter values:

\begin{table}[h]
\centering
\caption{Optimized parameters for the malware dynamics ODE model}
\label{tab:ode-params}
\begin{tabular}{clll}
\toprule
Parameter & Description & Value & Units \\
\midrule
$\alpha_0$ & Initial infection rate & 0.0501 & day$^{-1}$ \\
$\beta$ & Suppression coefficient & $10^{-4}$ & intensity$^{-1}$day$^{-1}$ \\
$\kappa$ & Feedback strength & 0.005 & day$^{-1}$ \\
$K$ & Carrying capacity & $10^5$ & intensity units \\
$p_{\text{decay}}$ & Temporal decay rate & 0.48 & dimensionless \\
\bottomrule
\end{tabular}
\end{table}

Note that time is measured in days, and intensity units represent the number of infected hosts observed per time bin. The dimensionless parameters ($p_{\text{decay}}$) are normalized to ensure numerical stability during integration.

Parameters were optimized by minimizing the mean squared error between model predictions and smoothed observations:

\begin{equation}
\mathcal{L}(\theta) = \frac{1}{N} \sum_{i=1}^{N} \left( M(t_i; \theta) - \eta_{\text{smooth}}(t_i) \right)^2
\end{equation}

\subsubsection{Numerical Implementation}

The ODE system exhibits numerical stiffness due to the logarithmic feedback term and quadratic nonlinearity. We implement the model in Julia using the \texttt{DifferentialEquations.jl} library \cite{rackauckas2017differentialequations} with the following specifications:

\begin{itemize}
    \item \textbf{Solver}: Rodas5, a 5th-order stiff-aware Rosenbrock method \cite{hairer1996solving}
    \item \textbf{Initial condition}: $M(0) = \max(\eta(0), 1.0)$ to ensure numerical stability
    \item \textbf{Non-negativity constraint}: $M(t) \geq 0$ enforced at each integration step
    \item \textbf{Integration tolerances}: $\texttt{abstol} = 10^{-6}$, $\texttt{reltol} = 10^{-6}$
\end{itemize}

The implementation algorithm is summarized below:

%\begin{algorithm}[h]
\begin{algorithm}[htbp]
\caption{Malware Dynamics ODE Implementation}
\label{alg:malware-ode}
\begin{algorithmic}[1]
\Function{malware\_ode!}{$du, u, p, t$}
    \State $M \gets \max(u[1], 0.0)$ \Comment{Enforce non-negativity}
    \State $\alpha, \beta, \kappa, K, p_{\text{decay}} \gets p$ \Comment{Unpack parameters}
    \State $\alpha_t \gets \alpha \cdot \exp(-p_{\text{decay}} \cdot t / t_{\max})$ \Comment{Time-dependent rate}
    \State $\text{growth} \gets \alpha_t \cdot M \cdot (1 - M/K)$
    \State $\text{external} \gets \eta_{\text{interp}}(t)$
    \State $\text{suppression} \gets -\beta \cdot M^2$
    \State $\text{feedback} \gets \kappa \cdot M \cdot \log(1 + M)$
    \State $du[1] \gets \text{growth} + \text{external} + \text{suppression} + \text{feedback}$
\EndFunction
\end{algorithmic}
\end{algorithm}

\subsection{Neural Ordinary Differential Equation (Neural ODE) Model}

The Neural ODE approach represents a fully data-driven paradigm that replaces traditional mechanistic modeling with end-to-end learning \cite{chen2018neural}. This method directly parameterizes the entire right-hand side of the differential equation with a neural network:

\begin{equation}
\frac{dM}{dt} = \mathcal{N}_{\psi}(M, t)
\label{eq:neural_ode}
\end{equation}

where $\mathcal{N}_{\psi}$ is a neural network parameterized by $\psi$ that directly learns the underlying dynamics from data without explicit mechanistic assumptions \cite{dupont2019augmented}. This approach offers maximum flexibility in capturing complex, nonlinear dynamics but sacrifices the interpretability provided by physics-based components \cite{kidger2022neural}.

\subsubsection{Network Architecture}

Our Neural ODE model employs a deeper architecture than the UDE counterpart to compensate for the absence of mechanistic structure. The increased capacity is necessary to learn both the fundamental dynamics and the complex interactions that the physics-based terms capture in hybrid models.

The network architecture consists of:
\begin{itemize}
    \item \textbf{Input}: 2-dimensional vector containing normalized malware intensity $\hat{M} = M/\max(\eta)$ and normalized time $\hat{t} = t/\max(t_{\text{data}})$
    \item \textbf{Hidden layers}: Two dense layers with 16 neurons each using ReLU activation functions
    \item \textbf{Output}: Single neuron producing the time derivative $dM/dt$
\end{itemize}

Mathematically, the network architecture can be expressed as:
\begin{equation}
\mathcal{N}_{\psi}(M, t) = \mathbf{W}_3^T \cdot \sigma(\mathbf{W}_2 \cdot \sigma(\mathbf{W}_1 \cdot [\hat{M}, \hat{t}] + \mathbf{b}_1) + \mathbf{b}_2) + b_3
\end{equation}

where $\psi = \{\mathbf{W}_1 \in \mathbb{R}^{16 \times 2}, \mathbf{b}_1 \in \mathbb{R}^{16}, \mathbf{W}_2 \in \mathbb{R}^{16 \times 16}, \mathbf{b}_2 \in \mathbb{R}^{16}, \mathbf{W}_3 \in \mathbb{R}^{16}, b_3 \in \mathbb{R}\}$ denotes all trainable parameters and $\sigma$ represents the ReLU activation function applied element-wise. The inclusion of time as an explicit input enables the network to learn time-dependent dynamics, which is critical for capturing the non-autonomous aspects of malware propagation where external factors vary over time.

\subsubsection{Training and Optimization}

We train the Neural ODE using a two-phase optimization strategy similar to the UDE approach, following established practices in neural differential equation optimization \cite{chen2018neural}:

\begin{enumerate}
    \item \textbf{Adam phase}: 300 iterations using the Adam optimizer \cite{kingma2015adam} with learning rate $5 \times 10^{-4}$ for broad parameter exploration and robust initial convergence
    \item \textbf{LBFGS phase}: 200 iterations for fine-tuning with L-BFGS, a quasi-Newton method that leverages second-order optimization for high-precision convergence
\end{enumerate}

The loss function remains consistent with previous models:
\begin{equation}
\mathcal{L}(\psi) = \frac{1}{N} \sum_{i=1}^{N} \left( M(t_i; \psi) - \eta_{\text{smooth}}(t_i) \right)^2
\end{equation}

\subsubsection{Implementation Details}

The Neural ODE implementation incorporates several numerical techniques to ensure stability during both training and inference, addressing common challenges in neural differential equation optimization:

\begin{algorithm}[htbp]
\caption{Neural ODE Implementation}
\label{alg:neural-ode}
\begin{algorithmic}[1]
\Function{neural\_ode!}{$du, u, p, t$}
    \State $p_{\text{restructured}} \gets \text{ComponentArray}(p, \text{getaxes}(p_{\text{flat}}))$ \Comment{Parameter handling}
    \State $M \gets \max(\min(u[1], 5 \cdot \max_\eta), 0.0)$ \Comment{State clamping for stability}
    \State $M_{\text{norm}} \gets M / \max_\eta$ \Comment{Normalize intensity}
    \State $t_{\text{norm}} \gets t / \max(t_{\text{data}})$ \Comment{Normalize time}
    \State $\text{nn\_input} \gets \text{reshape}([M_{\text{norm}}, t_{\text{norm}}], :, 1)$ \Comment{Create input tensor}
    \State $\text{nn\_out}, \text{nn\_state} \gets \mathcal{N}_\psi(\text{nn\_input}, p_{\text{restructured}}, \text{nn\_state})$
    \Comment{Forward pass}
    \State $du[1] \gets \text{clamp}(\text{nn\_out}[1], -1000.0, 1000.0)$ \Comment{Output bounding}
\EndFunction
\end{algorithmic}
\end{algorithm}

Key implementation considerations include:
\begin{itemize}
    \item \textbf{Parameter management}: ComponentArray structure for efficient gradient computation through the neural network, enabling seamless integration with automatic differentiation
    \item \textbf{Solver configuration}: Rodas5 stiff solver with adaptive tolerances $\text{abstol} = 10^{-3}$, $\text{reltol} = 10^{-3}$, optimized for neural differential equations
    \item \textbf{Numerical stability}: State clamping prevents unphysical values, input normalization ensures stable gradients, and output bounding prevents extreme derivatives that could destabilize integration
    \item \textbf{Error handling}: Graceful fallback mechanisms handle numerical difficulties that may arise during the iterative optimization process
\end{itemize}

This approach leverages the universal approximation capabilities of neural networks to potentially discover complex dynamics not captured by mechanistic models, albeit at the cost of reduced interpretability and increased computational requirements \cite{hornik1989multilayer}. The method is particularly valuable when the underlying physical mechanisms are poorly understood or when the system exhibits highly nonlinear behaviors that are difficult to express analytically.

\subsection{Universal Differential Equation (UDE) Model}

Universal Differential Equations (UDEs) combine mechanistic models with neural networks to capture complex dynamics that are difficult to express analytically. This hybrid approach preserves interpretability while improving flexibility by embedding data-driven components into known system structures \cite{rackauckas2020universal}.

In cybersecurity applications, UDEs are particularly effective for modeling adaptive feedback mechanisms and emergent network effects that arise from complex interactions between malware propagation, network topology, and defensive responses \cite{raissi2019physics}. The framework has demonstrated success across diverse scientific domains, from fluid dynamics \cite{sharma2023review} to epidemiological modeling \cite{dandekar2020machine}, making it well-suited for malware dynamics where both mechanistic understanding and adaptive learning are essential.

Building upon our classical ODE foundation, we enhance the model by replacing the  feedback term $\kappa M \log(1 + M)$ with a learnable neural network component $\mathcal{N}_{\phi}(M)$. This substitution yields a hybrid physics-informed model:

\begin{equation}
\frac{dM}{dt} = \underbrace{\alpha(t) M \left(1 - \frac{M}{K} \right)}_{\text{logistic growth}} + \underbrace{\eta(t)}_{\text{external forcing}} - \underbrace{\beta M^2}_{\text{quadratic suppression}} + \underbrace{\mathcal{N}_{\phi}(M)}_{\text{learned feedback}}
\label{eq:ude}
\end{equation}

This formulation preserves the interpretable mechanistic components while enabling the neural network to learn complex feedback mechanisms directly from data. The approach follows the scientific machine learning principle of embedding domain knowledge while maintaining flexibility for discovery \cite{willard2020integrating}.

\subsubsection{Neural Network Architecture}

The neural network \( \mathcal{N}_{\phi} \) is a lightweight feedforward model designed to represent unknown feedback mechanisms in the malware dynamics. It operates on normalized malware intensity values and outputs a learned correction term that complements the mechanistic structure of the differential equation.

The network architecture consists of:

\begin{itemize}
    \item \textbf{Input layer}: A single neuron accepting the normalized malware intensity \( \hat{M} = M/\max(M) \)
    \item \textbf{Hidden layer}: One hidden layer with 10 neurons using the ReLU activation function
    \item \textbf{Output layer}: A single neuron that outputs the learned feedback contribution
\end{itemize}

Mathematically, the neural network can be expressed as:
\begin{equation}
\mathcal{N}_{\phi}(M) = \mathbf{W}_2^T \cdot \sigma(\mathbf{W}_1 \cdot \hat{M} + \mathbf{b}_1) + b_2
\end{equation}

where \( \phi = \{\mathbf{W}_1, \mathbf{b}_1, \mathbf{W}_2, b_2\} \) denotes the trainable parameters of the network, and \( \sigma \) is the ReLU activation function.

\subsubsection{Training Methodology}

The UDE model is trained using a two-phase optimization strategy designed to balance exploration and precision:

\begin{enumerate}
    \item \textbf{Exploration phase}: We first perform 300 iterations using the Adam optimizer \cite{kingma2015adam} with a learning rate of \( 5 \times 10^{-4} \). This phase enables broad exploration of the parameter space and robust initial convergence.
    
    \item \textbf{Refinement phase}: Subsequently, we fine-tune the parameters using the L-BFGS optimizer for 200 iterations. This quasi-Newton method allows for high-precision convergence to a local minimum.
\end{enumerate}

The model is trained by minimizing the mean squared error (MSE) between the predicted malware intensity trajectory and the smoothed empirical observations:

\begin{equation}
\mathcal{L}(\phi, \theta) = \frac{1}{N} \sum_{i=1}^{N} \left( M(t_i; \phi, \theta) - \eta_{\text{smooth}}(t_i) \right)^2
\end{equation}

Here, \( \phi \) denotes the neural network parameters and \( \theta \) represents the set of ODE parameters. The smoothed empirical signal \( \eta_{\text{smooth}}(t) \) serves as the target trajectory for model fitting.

\subsubsection{Implementation Details}

Our UDE implementation incorporates several techniques to ensure stability during training:

%\begin{algorithm}[h]
\begin{algorithm}[htbp]

\caption{UDE Implementation}
\label{alg:ude-implementation}
\begin{algorithmic}[1]
\Function{ude\_hybrid!}{$du, u, p, t$}
    \State $M \gets \max(\min(u[1], 5 \cdot \max_\eta), 0.0)$ \Comment{State clamping}
    \State $\alpha, \beta, K, p_{\text{decay}} \gets \text{abs}(p.\text{ode})$ \Comment{Parameter positivity}
    \State $\alpha_t \gets \alpha \cdot \exp(-p_{\text{decay}} \cdot t / t_{\max})$
    \State $\text{growth} \gets \alpha_t \cdot M \cdot (1 - M / K)$
    \State $\text{suppression} \gets -\beta \cdot M^2$
    \State $\text{external} \gets \eta_{\text{interp}}(t)$
    \State $M_{\text{norm}} \gets M / \max_\eta$ \Comment{Input normalization}
    \State $\text{nn\_input} \gets \text{reshape}([M_{\text{norm}}], :, 1)$ \Comment{Prepare input tensor}
    \State $\text{nn\_out}, \text{nn\_state} \gets \mathcal{N}_\phi(\text{nn\_input}, p.\text{nn}, \text{nn\_state})$ \Comment{Forward pass}
    \State $\text{nn\_term} \gets \text{clamp}(\text{nn\_out}, -1000.0, 1000.0)$ \Comment{Output bounding}
    \State $du[1] \gets \text{growth} + \text{suppression} + \text{external} + \text{nn\_term}$
\EndFunction
\end{algorithmic}
\end{algorithm}

Key implementation details include:
\begin{itemize}
    \item \textbf{Parameter structure}: Combined ComponentArray with ODE parameters and neural network weights
    \item \textbf{Gradient computation}: Automatic differentiation through the ODE solver
    \item \textbf{Numerical stability}: Input normalization, output clamping, and parameter positivity constraints
\end{itemize}

\section{Results}
The evaluation emphasizes the Code Red dataset for its thorough coverage and data quality; the UDE framework aims to generalize to various malware propagation scenarios. The Code Red outbreak provides an ideal test case as it represents a well-documented, large-scale malware event with complete temporal dynamics.

We evaluate the three modeling approaches, ODE, UDE, and Neural ODE, across multiple dimensions to assess their capabilities in capturing malware dynamics.

Performance is primarily measured using Root Mean Square Error (RMSE), defined as:
\begin{equation}
\text{RMSE} = \sqrt{\frac{1}{N}\sum_{i=1}^{N}(y_i - \hat{y}_i)^2}
\end{equation}
where $y_i$ represents observed values and $\hat{y}_i$ represents model predictions.

\subsection{Model Performance Comparison}

\begin{figure}[!htbp]
    \centering
    \includegraphics[width=0.7\linewidth]{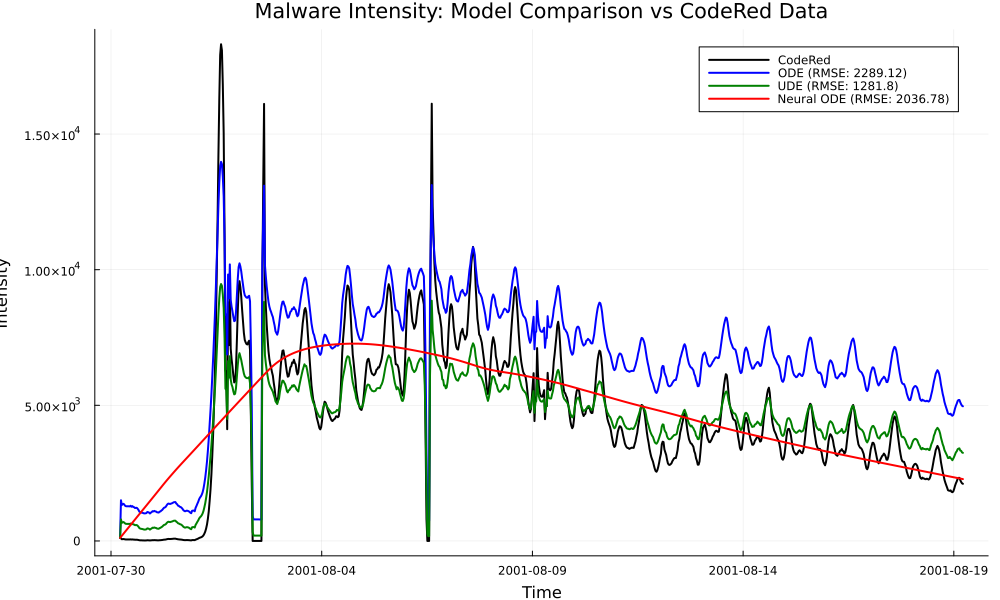}
    \caption{Comparison of model predictions against malware intensity data. The UDE model (green) achieves the lowest RMSE of 1281.8, outperforming both the traditional ODE model (blue, RMSE: 2289.12) and Neural ODE (red, RMSE: 2036.78).}
    \label{fig:model_comparison}
\end{figure}

Figure~\ref{fig:model_comparison} illustrates the fitting performance of all three modeling approaches against the Code Red outbreak data. The UDE approach demonstrates improved performance (RMSE: 1281.8) compared to both the physics-based ODE model (RMSE: 2289.12) and the fully data-driven Neural ODE (RMSE: 2036.78). Table~\ref{tab:performance_metrics} provides a comprehensive comparison of performance metrics.

\begin{table}[!htbp]
\centering
\caption{Performance metrics for malware dynamics models}
\label{tab:performance_metrics}
\begin{tabular}{lcccc}
\toprule
\textbf{Model} & \textbf{RMSE} & \textbf{MAE} & \textbf{MAPE (\%)} & \textbf{Correlation} \\
\midrule
ODE & 2289.12 & 2174.75 & 1503.79 & 0.948 \\
Neural ODE & 2036.78 & 1304.02 & 9368.04 & 0.687 \\
UDE & 1281.80 & 883.90 & 449.32 & 0.946 \\

\bottomrule
\end{tabular}
\end{table}

The UDE model achieves a 44.0\% reduction in RMSE compared to the traditional ODE approach while maintaining high correlation (0.946). The pure Neural ODE model, despite its flexibility, performs worse than the hybrid approach, suggesting that completely abandoning the physics-based structure reduces model effectiveness.

\subsection{Component Contribution Analysis}

\begin{figure}[!htbp]
    \centering
    \includegraphics[width=0.7\linewidth]{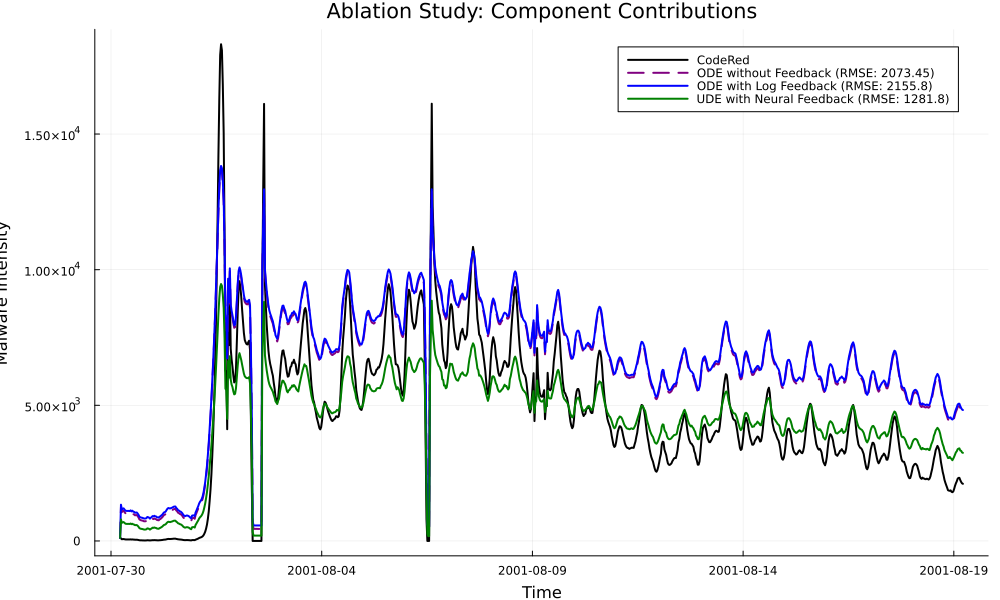}
    \caption{Ablation study comparing models with progressively more sophisticated feedback mechanisms. The UDE model with neural feedback (green, RMSE: 1281.8) substantially outperforms both the ODE without feedback (purple, RMSE: 2073.45) and ODE with standard logarithmic feedback (blue, RMSE: 2155.8).}
    \label{fig:ablation}
\end{figure}

To understand which components drive the UDE model's  performance, we conducted an ablation study by systematically evaluating different feedback mechanisms (Figure~\ref{fig:ablation}). This analysis isolates the contribution of each modeling component by comparing three variants:

\begin{enumerate}
    \item \textbf{No feedback model}: ODE without any feedback term ($\kappa=0$), RMSE: 2073.45
    \item \textbf{Analytical feedback model}: ODE with standard logarithmic feedback ($\kappa M \log(1+M)$), RMSE: 2155.8  
    \item \textbf{Learned feedback model}: UDE with neural network feedback, RMSE: 1281.8
\end{enumerate}

The analysis reveals several key insights:

\begin{itemize}
    \item The baseline ODE without feedback captures fundamental growth dynamics but lacks adaptive capabilities to handle complex outbreak patterns
    \item The standard logarithmic feedback term slightly degrades performance (2073.45 → 2155.8 RMSE), indicating that this particular analytical form may not adequately represent the true feedback mechanisms present in real malware propagation
    \item Replacing the fixed analytical feedback with a learnable neural network component dramatically improves performance, achieving a 40.5\% reduction in RMSE compared to the analytical feedback model
\end{itemize}

This ablation study demonstrates that while physics-based structure provides valuable inductive bias for capturing fundamental dynamics, the neural network component is crucial for learning complex, non-linear feedback mechanisms that analytical forms fail to capture. The improved performance of the learned feedback approach suggests that real-world malware dynamics involve feedback processes that are more complex than simple logarithmic relationships.

\subsection{Forecasting Capabilities}

\begin{figure}[!htbp]
    \centering
    \includegraphics[width=0.7\linewidth]{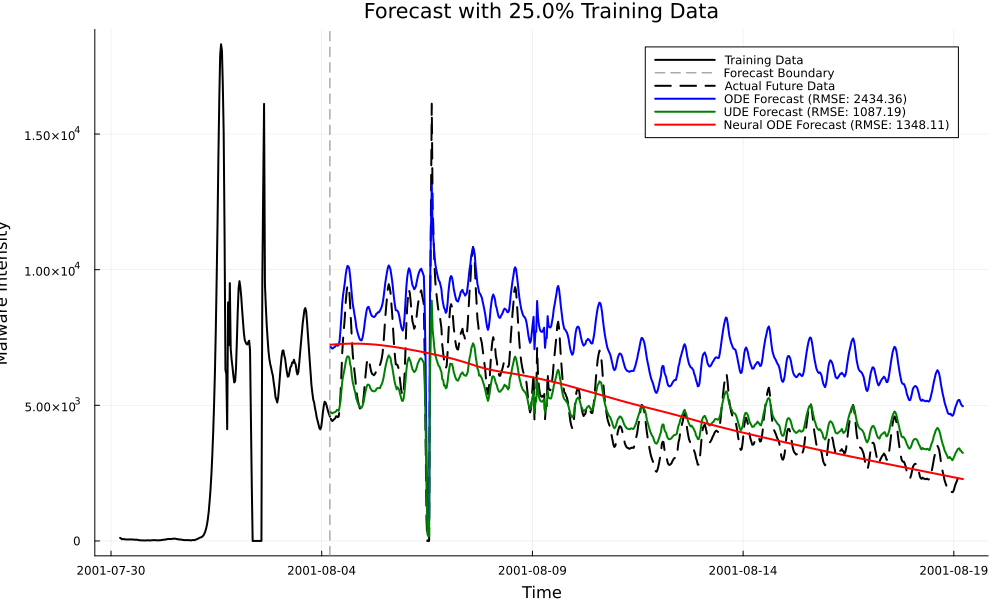}
    \caption{Forecasting performance comparison using only 25\% of data for training. The UDE model (green, RMSE: 1087.19) maintains accurate predictions even with limited training data, outperforming ODE (blue, RMSE: 2434.36) and Neural ODE (red, RMSE: 1348.11).}
    \label{fig:forecast_25}
\end{figure}

We assessed each model's ability to forecast beyond the training data by training on different subsets of the time series. Figure~\ref{fig:forecast_25} shows the forecast performance with just 25\% training data, demonstrating UDE's remarkable ability to generalize from limited observations. The vertical dashed line indicates the boundary between training and forecasting periods.

The UDE model maintains strong forecasting performance even with minimal training data (RMSE: 1087.19 at 25\%), while both the traditional ODE model (RMSE: 2434.36) and Neural ODE (RMSE: 1348.11) show significantly higher errors. Notably, the UDE model trained on only 25\% of the data outperforms the Neural ODE by 19.4\% and the traditional ODE by 55.3\%. This enhanced forecast stability with limited data is a crucial advantage for cybersecurity applications, where early detection and response with minimal observations is essential for effective malware containment.

\subsection{Robustness to Noise}

\begin{figure}[!htbp]
    \centering
    \includegraphics[width=0.7\linewidth]{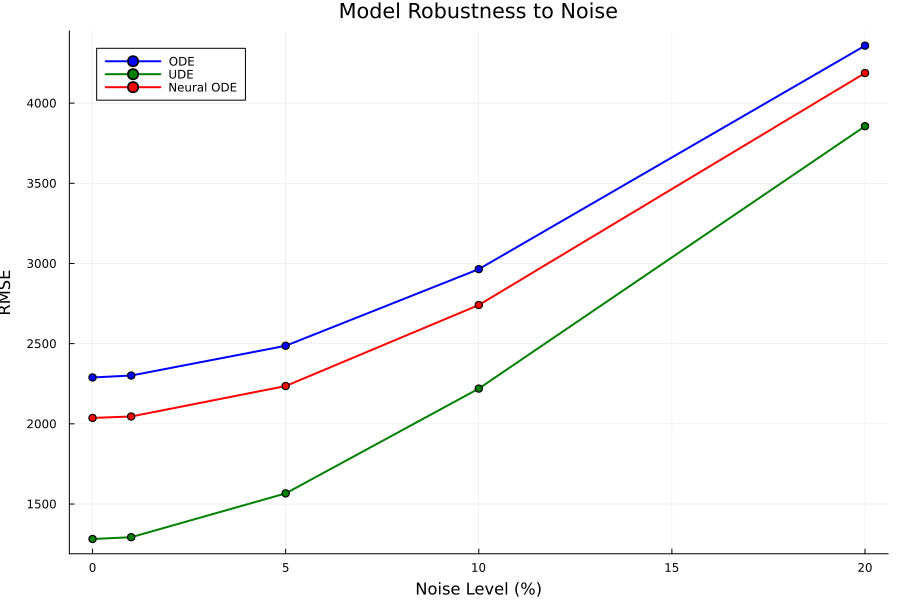}
    \caption{Model performance degradation with increasing noise levels. The UDE model (green) consistently demonstrates greater robustness to noise compared to both ODE (blue) and Neural ODE (red) approaches.}
    \label{fig:noise_sensitivity}
\end{figure}

Practical malware monitoring involves substantial measurement noise due to incomplete monitoring coverage, sampling limitations, and reporting inconsistencies, making model robustness critical for real-world deployment. We evaluated each model's performance across varying noise levels (0\%, 1\%, 5\%, 10\%, and 20\%), as shown in Figure~\ref{fig:noise_sensitivity}.

The UDE model demonstrates better noise robustness across all noise levels, maintaining the lowest RMSE values throughout the tested range. At 10\% noise, the UDE model (RMSE: 2219.82) substantially outperforms both the ODE (RMSE: 2964.69) and Neural ODE (RMSE: 2741.06) approaches, representing performance advantages of 25.1\% and 19.0\% respectively. Even at the highest tested noise level of 20\%, the UDE model maintains its performance advantage over both alternatives.

This enhanced robustness likely stems from the UDE's hybrid architecture: the physics-based components provide stability against random variations by enforcing fundamental constraints, while the neural network component maintains flexibility to adapt to the underlying signal patterns despite noise contamination. This combination proves particularly valuable for cybersecurity applications where data quality can vary significantly across different monitoring infrastructures.

\subsection{Symbolic Recovery and Interpretability}

\begin{figure}[!htbp]
    \centering
    \includegraphics[width=0.7\linewidth]{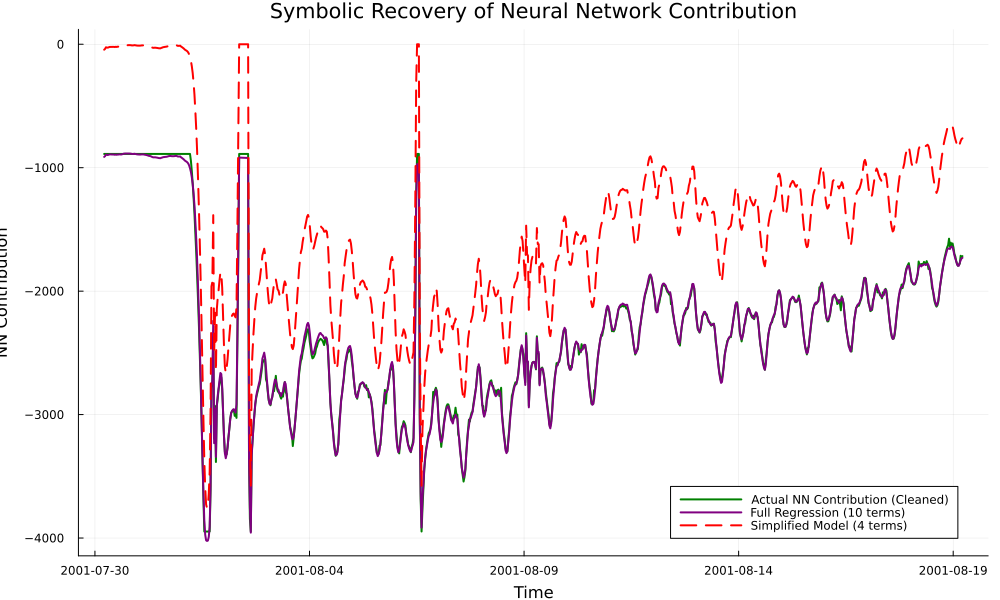}
    \caption{Symbolic recovery of the neural network contribution. The actual neural network output (green) can be approximated by a full regression model with 10 terms (purple) or a simplified model with just five terms (red dashed). The predominantly negative contribution demonstrates the neural network acting as a suppression mechanism.}
    \label{fig:symbolic_recovery}
\end{figure}

A key limitation of neural network approaches is their black-box nature, which hinders interpretability and limits trust in critical applications \cite{rudin2019stop}. To address this challenge, we applied symbolic regression techniques to recover explicit mathematical expressions approximating the neural network's contribution (Figure~\ref{fig:symbolic_recovery}). Following established approaches in discovering governing equations from data \cite{brunton2016discovering, schmidt2009distilling}, we employed regularized ridge regression \cite{tibshirani1996regression} to identify parsimonious mathematical expressions that capture the essential dynamics while maintaining interpretability \cite{ribeiro2016should}. We generated candidate symbolic terms, including polynomial, logarithmic, and rational functions, and then applied ridge regression with a regularization parameter of $\lambda=1.0$ to identify the most significant contributors while preventing overfitting.
Using ridge regression with optimal regularization parameters, we identified a concise 5-term approximation that balances accuracy with interpretability:

\begin{equation}
\mathcal{N}(M) \approx -2608.692 \cdot \frac{M}{1+M} - 2459.9124 \cdot \log(1+M) - 2113.3055 \cdot M + 1366.5024 \cdot M^{2} + 831.707 \cdot M \cdot \log(1+M)
\label{eq:symbolic}
\end{equation}

This symbolic representation reveals several critical insights:

\begin{enumerate}
    \item The neural network primarily functions as a \textit{suppression mechanism}, with an overwhelmingly negative contribution that increases with malware intensity
    \item This suggests traditional epidemic models systematically overestimate malware spread in real-world networks
    \item The recovered formula includes both negative (suppressive) and positive (amplifying) terms, reflecting a complex balance of competing forces in malware propagation dynamics
\end{enumerate}

\subsubsection{Cybersecurity Interpretation of Mathematical Terms}

\begin{figure}[!htbp]
    \centering
    \includegraphics[width=0.7\linewidth]{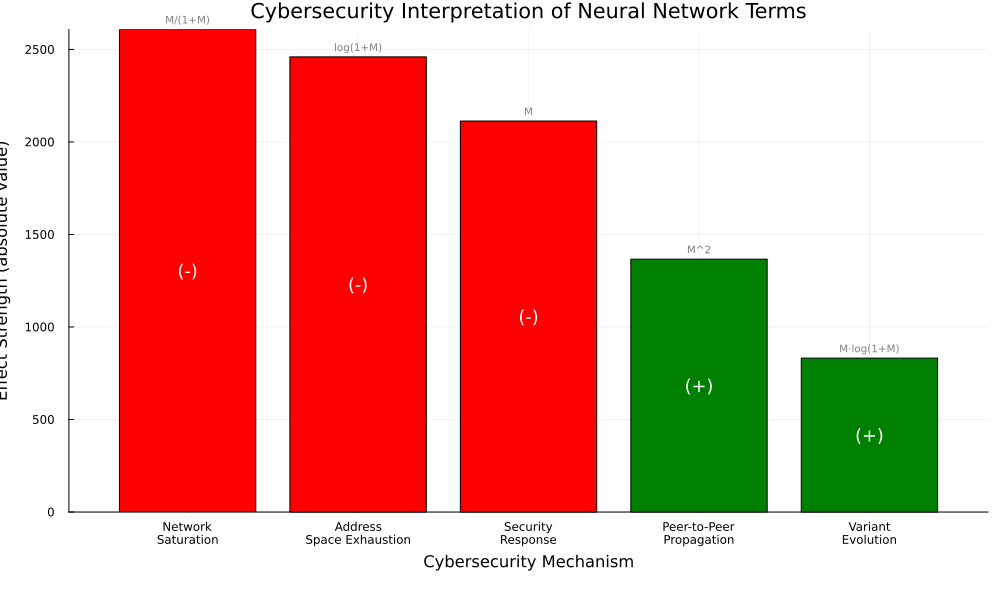}
    \caption{Cybersecurity interpretation of recovered symbolic terms. The height of each bar represents the absolute coefficient value, while the sign (+ or -) indicates whether the term accelerates or suppresses the spread of malware. The largest terms implement suppression mechanisms that correspond to real-world network effects.}
    \label{fig:term_mapping}
\end{figure}

To translate these mathematical findings into actionable security insights, we analyzed how each term corresponds to known cybersecurity phenomena (Figure~\ref{fig:term_mapping}). This analysis bridges the gap between mathematical modeling and practical cybersecurity by mapping abstract mathematical terms to concrete network mechanisms observed in real-world malware incidents.

Each term in Equation~\ref{eq:symbolic} corresponds to a specific cybersecurity mechanism:

\begin{itemize}
    \item $\frac{M}{1+M}$ term (coefficient: -2608.692) → \textbf{Network saturation effects}, where traffic congestion and resource contention limit further spread as network infrastructure becomes overwhelmed \cite{staniford2002practical}.
    \item $\log(1+M)$ term (coefficient: -2459.9124) → \textbf{Address space exhaustion}, representing diminishing returns in finding new vulnerable hosts as larger portions of the attack surface become compromised.
    \item $M$ term (coefficient: -2113.3055) → \textbf{Security response mechanisms}, including automated defenses and human-driven countermeasures deployed proportionally to observed infection levels \cite{zou2003monitoring}.
    \item $M^2$ term (coefficient: +1366.5024) → \textbf{Peer-to-peer propagation effects}, enabling quadratic growth through direct host-to-host transmission and lateral movement within compromised networks.
    \item $M \cdot \log(1+M)$ term (coefficient: +831.707) → \textbf{Variant evolution and adaptation}, capturing how successful malware spawns variants and develops techniques to overcome defensive measures.

\end{itemize}

The dominant negative contributions reveal that real-world malware faces substantial limiting factors beyond those captured in traditional epidemic models. These mechanisms include network congestion, security responses, and address space constraints that actively suppress unconstrained growth. Simultaneously, the positive terms reflect how successful malware can partially overcome these limitations through adaptation and efficient propagation techniques.

This interpretation aligns with cybersecurity practitioners' understanding of real-world malware dynamics, providing actionable insights for effective defense strategies. The analysis suggests that effective countermeasures should focus on enhancing the natural suppression mechanisms, such as network segmentation to amplify saturation effects, rapid patch deployment to accelerate security responses, and threat intelligence sharing to improve collective defense capabilities.

\section{Comparative Analysis of Modeling Approaches}

Having evaluated each modeling approach individually across multiple dimensions, we now provide a systematic comparison to identify their relative strengths, limitations, and optimal use cases. This analysis synthesizes our experimental findings to provide practical guidance for selecting suitable modeling strategies in various cybersecurity scenarios.

\begin{table}[!htbp]
\centering
\caption{Comprehensive comparison of modeling approaches across evaluation dimensions}
\label{tab:comprehensive_comparison}
\begin{tabular}{lccc}
\toprule
\textbf{Evaluation Dimension} & \textbf{ODE} & \textbf{Neural ODE} & \textbf{UDE} \\
\midrule
\textbf{Overall RMSE (Smoothed)} & 2289.12 & 2036.78 & \textbf{1281.80} \\
\textbf{Forecasting (25\% training)} & 2434.36 & 1348.11 & \textbf{1087.19} \\
\textbf{Forecasting (50\% training)} & 2696.06 & 912.39 & \textbf{697.29} \\
\textbf{Forecasting (75\% training)} & 2783.45 & \textbf{661.86} & 777.35 \\
\textbf{Noise Robustness (10\%)} & 2964.69 & 2741.06 & \textbf{2219.82} \\
\textbf{Noise Robustness (20\%)} & 4358.34 & 4187.65 & \textbf{3856.18} \\
\textbf{Correlation} & \textbf{0.948} & 0.687 & 0.946 \\
\textbf{Neural Network Parameters} & \textbf{0} & 337 & 31 \\
\textbf{Interpretability} & \textbf{High} & Low & \textbf{High} \\
\bottomrule
\end{tabular}
\end{table}

\subsection{Performance and Data Efficiency}

The UDE approach demonstrates improved performance across most accuracy metrics, achieving a 44.0\% improvement over traditional ODEs in overall fitting performance. However, analysis of forecasting capabilities reveals important data dependency patterns: while UDEs maintain consistent performance across all training data sizes (1087.19 to 777.35 RMSE), Neural ODEs show dramatic improvement with increased training data, achieving competitive results with abundant data (661.86 RMSE at 75\% training) but poor performance with limited data (1348.11 RMSE at 25% training).

This data dependency has critical implications for cybersecurity applications. UDEs trained on just 25\% of data (RMSE: 1087.19) significantly outperform Neural ODEs with the same limited training data (RMSE: 1348.11), making UDEs particularly valuable for early outbreak detection scenarios where historical data is scarce. However, for forensic analysis with complete datasets, Neural ODEs can achieve competitive forecasting performance.

\subsection{Robustness and Interpretability}

UDEs demonstrate improved performance across all noise levels, consistently maintaining the lowest RMSE values as shown in Table~\ref{tab:comprehensive_comparison}. At high noise levels (20\%), UDEs achieve approximately 8\% better performance than Neural ODEs (3856.18 vs 4187.65 RMSE) and 11\% better than traditional ODEs (3856.18 vs 4358.34 RMSE).

However, analysis of relative performance degradation reveals an important trade-off: while UDE maintains the best absolute performance, it experiences the highest proportional degradation from baseline (201\% increase at 20\% noise), compared to traditional ODE (90\% increase) and Neural ODE (106\% increase). This suggests that UDE's enhanced baseline performance comes with increased sensitivity to noise corruption, though it still outperforms alternatives in absolute terms across all tested noise levels.

UDEs preserve interpretability through symbolic recovery, offering an optimal balance between performance and explainability. Traditional ODEs offer full interpretability but limited performance, whereas Neural ODEs sacrifice interpretability for flexibility, as evidenced by substantially lower correlation scores (0.687) compared to both UDEs (0.946) and traditional ODEs (0.948).

\subsection{Component Contribution Analysis}

To understand the UDE's performance, we conducted an ablation study examining the contribution of different feedback mechanisms. This analysis reveals critical insights about the effectiveness of learned versus analytical feedback components.

\begin{table}[!htbp]
\centering
\caption{Component contribution analysis through ablation study}
\label{tab:ablation_results}
\begin{tabular}{lcc}
\toprule
\textbf{Model Configuration} & \textbf{RMSE} & \textbf{Improvement} \\
\midrule
ODE without Feedback & 2073.45 & Baseline \\
ODE with Log Feedback ($\kappa M \log(1+M)$) & 2155.80 & -3.97\% \\
UDE with Neural Feedback & \textbf{1281.80} & \textbf{+40.54\%} \\
\bottomrule
\end{tabular}
\end{table}

The results demonstrate that traditional analytical feedback mechanisms not only fail to improve performance but degrade it by 3.97\%. This suggests that conventional epidemiological assumptions about logarithmic feedback in malware dynamics \cite{kephart1991directed, zou2002code} are inappropriate for real-world network environments where complex topology and infrastructure effects dominate \cite{pastor2001dynamical}. In contrast, the neural network component achieves a remarkable 40.54\% improvement over the baseline physics-only model, highlighting the critical importance of learned feedback mechanisms in capturing the complex dynamics of malware propagation.

This finding has profound implications: it indicates that the neural network component has discovered feedback mechanisms that are fundamentally different from traditional analytical forms, justifying the hybrid UDE approach and explaining why pure Neural ODEs, despite their flexibility, cannot match UDE performance due to the lack of physics-based structural constraints. Significantly, the UDE achieves this enhanced performance with significantly fewer trainable neural network parameters (31) compared to the Neural ODE (337 parameters), demonstrating that incorporating physics-based structure enables the neural component to focus on learning only the residual dynamics that analytical models cannot capture, rather than having to learn the entire system behavior from scratch.

\subsection{Practical Recommendations and Cybersecurity Applications}

Based on our comprehensive evaluation across forecasting, noise robustness, and component contribution analyses, we provide specific guidance for selecting appropriate modeling approaches in different cybersecurity scenarios:

\textbf{Recommended Use Cases by Scenario:}

\begin{itemize}
    \item \textbf{Early Warning Systems (Limited Data)}: UDE is recommended due to its improved performance with minimal training data (1087.19 RMSE vs 1348.11 for Neural ODE at 25\% training), making it ideal for detecting emerging threats with limited historical information \cite{zou2003monitoring}.
    
    \item \textbf{Forensic Analysis (Complete Datasets)}: Neural ODE performs best with abundant training data (661.86 RMSE at 75\% training), making it suitable for post-incident analysis where complete outbreak data is available.
    
    \item \textbf{Real-time Monitoring (Noisy Environments)}: UDE maintains the most consistent performance across all noise levels (3856.18 RMSE at 20\% noise vs 4187.65 for Neural ODE), which is critical for operational environments with data quality issues.
    
    \item \textbf{Resource-Constrained Environments}: UDE offers the best balance of performance and efficiency, achieving improved accuracy with 11 times fewer neural network parameters (31 vs. 337) than Neural ODE. Traditional ODE should only be considered when neural network training is completely infeasible, accepting significant performance trade-offs.
    
    \item \textbf{Interpretable Security Analysis}: UDE provides an optimal combination of enhanced performance and interpretability through symbolic recovery, enabling security teams to understand malware spread mechanisms and inform targeted defense strategies.
\end{itemize}

UDE provides an optimal combination of enhanced performance and interpretability through symbolic recovery, enabling security teams to understand malware spread mechanisms and inform targeted defense strategies. Our findings highlight that learned feedback mechanisms capture suppression and amplification effects that traditional models often overlook. This enables accurate threat assessment, better resource allocation, and informed policy development, reducing both under-response to real threats and over-response to false alarms, thereby facilitating more efficient cyber defense planning.

\section{Conclusion}

This work demonstrates that Universal Differential Equations (UDEs) offer a robust framework for modeling malware propagation, combining the interpretability of physics-based models with the adaptability of neural networks. Through systematic evaluation across forecasting accuracy, noise robustness, and data efficiency, we show that hybrid physics-neural approaches consistently outperform both traditional analytical models and purely neural methods.

Our symbolic recovery analysis reveals a key insight: real-world malware propagation exhibits suppression mechanisms, such as network saturation, security response, and variant evolution, that are absent from conventional epidemiological frameworks. By translating learned neural feedback into interpretable mathematical expressions, we bridge the gap between abstract modeling and actionable cybersecurity understanding.

These findings have immediate implications for cybersecurity practice. The UDE framework’s ability to perform well with limited data makes it especially suited for early warning systems during emerging outbreaks. Moreover, its interpretability empowers analysts to understand and leverage inherent network constraints for more effective intervention. The demonstrated inadequacy of traditional epidemiological feedback models suggests that malware dynamics require domain-specific formulations rather than direct analogies to biological epidemics.

While our results demonstrate the effectiveness of the UDE approach, we acknowledge some limitations in this initial study. Our evaluation relies on the Code Red worm dataset. The analysis focuses on initial outbreak dynamics and does not include statistical significance testing or confidence intervals for the reported performance improvements. Additionally, the optimal UDE parameters identified for Code Red may require retraining for different malware families, which limits their immediate applicability to emerging threats without further validation across diverse malware types.

Future work will focus on validating the UDE framework on modern malware datasets, including ransomware and botnets, to confirm the generalizability of our findings. Key priorities include developing methods to transfer UDE parameters across different malware families and testing the framework's real-time performance for early warning systems. Additionally, incorporating actual network topology data could further enhance the model's accuracy and practical applicability for cybersecurity operations.

\section*{Acknowledgments}

We gratefully acknowledge the assistance of several large language models (LLMs) in preparing this manuscript. Anthropic's Claude 3.5 Sonnet was utilized for coding support with the Julia experiments and debugging assistance. Google's Gemini 2.0 Flash provided grammar correction support, while Anthropic's Claude 4 Sonnet helped resolve LaTeX formatting issues. All models were accessed via their publicly available interfaces. We thank Anthropic and Google for making these powerful tools available to support academic research.

\bibliographystyle{ieeetr}
\bibliography{references}  

\end{document}